\documentclass[runningheads]{llncs}
\usepackage[T1]{fontenc}
\usepackage{graphicx}
\usepackage{hyperref}
\usepackage{tikz}
\usepackage{array}
\usepackage{multirow}
\usepackage{pgfplots}
\usepackage{tikz}
\begin{document}
\title{Text2Gender: A Deep Learning Architecture for Analysis of Blogger's Age and Gender}
%
%
\author{Vishesh Thakur\ \and
Aneesh Tickoo}
\authorrunning{Vishesh Thakur et al.}
%
\institute{Department of Computer Science and Engineering, Indian Institute of Technology Bhilai, Chhattisgarh, India\\
\email{\{visheshthakur,aneeshtickoo\}@iitbhilai.ac.in}}
\maketitle              
\begin{abstract}
\begin{quote}
Deep learning techniques have gained a lot of traction in the field of NLP research. The aim of this paper is to predict the age and gender of an individual by inspecting their written text. We propose a supervised BERT-based classification technique in order to predict the age and gender of bloggers. The dataset used contains 681284 rows of data, with the information of the blogger's age, gender, and text of the blog written by them. We compare our algorithm to previous works in the same domain and achieve a better accuracy and F1 score. The accuracy reported for the prediction of age group was 84.2\%, while the accuracy for the prediction of gender was 86.32\%. This study relies on the raw capabilities of BERT to predict the classes of textual data efficiently. This paper shows promising capability in predicting the demographics of the author with high accuracy and can have wide applicability across multiple domains. 
\end{quote}
\end{abstract}

\section{Introduction}
Gender estimation based on textual content has been a common area of research in the field of natural language processing. Blogs, in particular, provide a rich source of information for studying gender-related patterns, as authors often express their thoughts, opinions, and experiences through their writings. The ability to accurately estimate the gender of blog authors can have significant implications in various domains such as marketing, social sciences, and online content personalization. 

In this study, we focus on estimating the gender of blog authors based on the content they write. Our approach leverages the power of advanced language models, specifically BERT (Bidirectional Encoder Representations from Transformers)\cite{devlin2019bert}, for encoding the textual information present in the blogs. BERT has gained immense popularity for its ability to capture deep contextual representations of text, making it well-suited for various natural language processing tasks.

To build our gender estimation model, we employ a supervised learning approach. We utilize a large dataset of blogs where the gender of the authors is known. The blogs are preprocessed and encoded using BERT, which generates high-dimensional representations capturing the semantic and syntactic properties of the text. These encoded representations serve as input features for training a machine learning model, with gender as the categorical target variable.

One key aspect of our work is the novel application of advanced sentence models like BERT for gender estimation in text. While previous studies\cite{mukherjee}\cite{abdul}\cite{inbook} have explored gender estimation using traditional feature engineering techniques\cite{Goswami_Sarkar_Rustagi_2009} and simpler models, the utilization of BERT opens up new possibilities for capturing more nuanced patterns and improving the accuracy of gender predictions.

By employing BERT as the encoding backbone and training our model on a large dataset, we aim to achieve robust and reliable gender estimation results. Through rigorous evaluation and analysis, we will assess the performance of our model and compare it with existing approaches in the field. The insights gained from this research can contribute to a deeper understanding of the relationship between language use and gender, furthering our knowledge in the realm of computational gender studies.

In summary, this study presents a novel approach to estimating the gender of blog authors based on their writings. By leveraging the power of BERT, we aim to achieve accurate gender predictions and explore the potential of advanced sentence models in the domain of Text2Age. Our work contributes to the growing body of research in gender estimation and showcases the value of state-of-the-art language models in addressing challenging natural language processing tasks.

\section{Related Work}
Previous research has explored the usage of language patterns by different social groups, although limited by the availability of annotated data. The advent of the blogosphere and its accessibility for data collection has greatly facilitated research in this area. Several studies have examined the effects of bloggers' age and gender on language usage in weblogs \cite{inproceedings}; \cite{burger}; Yan, 2006; \cite{2003}; \cite{Yan2006GenderCO}; \cite{scott}).

However, a comprehensive analysis of bloggers' age and gender based on more advanced language models like BERT is lacking in the existing literature. Previous studies have primarily relied on keyword analysis, parts of speech, and other grammatical constructs for gender estimation and age determination. While some progress has been made in gender estimation \cite{inproceedings}, less attention has been given to age determination.

For instance, \cite{pennebaker} and Pennebaker and \cite{pandstone} examined age-related variations in language use, but their approaches were not specifically tailored to blog data. Burger and Henderson (2006) also explored age-related patterns, but a more comprehensive analysis using advanced language models is still needed.

Additionally, gender estimation has received more attention in previous research. Koppel, Argamon, and Shimoni (2003) estimated authors' gender using the British National Corpus text, while Schler et al. (2006) achieved 80\% accuracy in classifying authors' gender by analyzing function words and part-of-speech. Their findings indicated that female authors tend to use pronouns with high frequency, while male authors tend to use numerals and representation-related numbers more frequently.

However, these previous studies were limited in their approaches and did not leverage advanced sentence models like BERT, which have shown promising results in various natural language processing tasks. Our research aims to address this gap by employing BERT for gender estimation and exploring its potential in determining the age of blog authors based on their writings.

By utilizing advanced language models and training our model on a large annotated dataset, we aim to achieve higher accuracy in gender estimation and age determination compared to previous studies. The incorporation of BERT allows us to capture deeper contextual representations and uncover more nuanced patterns in language use.

In summary, while previous research has examined the relationship between language patterns and gender or age in the context of weblogs\cite{scott}, our study contributes to the existing literature by employing advanced language models like BERT. This enables us to enhance the accuracy of gender estimation and explore the feasibility of determining the age of blog authors. Our work aims to advance the field of computational gender studies and provide valuable insights into the analysis of bloggers' gender and age based on their written content.

\section{Dataset and Computation Environment}
The dataset\cite{inproceedings} we used for this study is called the 'Blog Authorship Corpus'. The Blog Authorship Corpus is a collection of posts written by 19,320 bloggers, which were obtained from blogger.com in August 2004. The corpus comprises a total of 681,288 posts and more than 140 million words, which translates to roughly 35 posts and 7250 words per person. Each blog is presented in a separate file and is identified by a unique blogger ID, as well as the blogger's self-reported gender, age, industry, and astrological sign. While all bloggers' genders and ages are labeled, industry and sign information is unknown for some.
The corpus is divided into three age groups: 8240 "10s" blogs (ages 13-17), 8086 "20s" blogs (ages 23-27), and 2994 "30s" blogs (ages 33-47), with an equal number of male and female bloggers in each group. Additionally, each blog in the corpus contains at least 200 occurrences of common English words. The corpus is devoid of all formatting except for two exceptions, i.e., individual posts within a single blogger are separated by the date of the subsequent post, and links within a post are marked by the label URL link.
\begin{center}
\begin{tabular}{||c c||} 
 \hline
 Age Group & Number of Subjects \\ [0.5ex] 
 \hline\hline
 10s (13 to 17) & 8240  \\ 
 \hline
 20s (23 to 27) & 8086 \\
 \hline
 30s (33 to 47) & 2994 \\[1ex] 
 \hline
\end{tabular}
\end{center}
Moreover, the population of Males and Females is equally distributed among all the classes. \break
Now let's have a look at the final data that is presented in the form of entire dataset. We have seen above that which age group has how many unique participants or subjects. Now let's have a look at what is the number for the total dataset as although the participants are unique, the texts are not. This essentially means that a single author may write more than one, or multiple pieces of text, which is fairly understood. Let's have a look at the data about the demographics of all the texts. First, let's see about the data distribution of age groups.
\begin{center}
\begin{tabular}{||c c||} 
 \hline
 Age Group & Number of Subjects \\ [0.5ex] 
 \hline\hline
 10s (13 to 17) & 236024  \\ 
 \hline
 20s (23 to 27) & 321447 \\
 \hline
 30s (33 to 47) & 98056 \\[1ex] 
 \hline
\end{tabular}
\end{center}
The distribution of data among the two genders is as below.
\begin{center}
\begin{tabular}{||c c||} 
 \hline
 Gender & Number of Subjects \\ [0.5ex] 
 \hline\hline
Male & 345193  \\ 
 \hline
 Female & 336091 \\[1ex] 
 \hline
\end{tabular}
\end{center}
The data is almost fairly distributed among the male and female populations as per the final dataset. \break
To feed the data into the network, we had to break the data into training and testing sets. We did a 4:1 split and the distribution of our data into the training and testing set is as follows.
\begin{center}
\begin{tabular}{||c c||} 
 \hline
 Training Set & 510964 \\ [0.5ex] 
 \hline\hline
 Testing Set & 170320 \\[1ex] 
 \hline
\end{tabular}
\end{center}
The computing system used was a commercial laptop with 8GB RAM, 2 GB NVIDIA GeForce MTX250 graphics card, and Intel i7 10th generation processor.

\section{Preprocessing}
Preprocessing of data played an important role in improving the accuracy of the data. For the prediction of age group, accuracy without preprocessing was 67.6\%, while after preprocessing it improved to 84.2\%, while for prediction of gender, before preprocessing it was 70.6\%, and after preprocessing became 86.32\%. \\

\usetikzlibrary{shapes.geometric, arrows}

\tikzstyle{startstop} = [rectangle, rounded corners, 
minimum width=3cm, 
minimum height=1cm,
text centered, 
draw=black, 
fill=white!30]

\tikzstyle{io} = [trapezium, 
trapezium stretches=true, 
trapezium left angle=70, 
trapezium right angle=110, 
minimum width=3cm, 
minimum height=1cm, text centered, 
draw=black, fill=blue!30]

\tikzstyle{process} = [rectangle, 
minimum width=3cm, 
minimum height=1cm, 
text centered, 
text width=3cm, 
draw=black, 
fill=orange!30]

\tikzstyle{decision} = [diamond, 
minimum width=3cm, 
minimum height=1cm, 
text centered, 
draw=black, 
fill=green!30]
\tikzstyle{arrow} = [thick,->,>=stealth]

\begin{tikzpicture}[node distance=2cm]

\node (start) [startstop] {Input blog text};
\node (in1) [startstop, below of=start] {Removing unnecessary characters (hashtags, emojis etc.)};
\node (in2) [startstop, below of=in1] {Stop words removal};
\node (pro1) [startstop, below of=in2] {Normalization of text and lemmatization};
\draw [arrow] (start) -- (in1);
\draw [arrow] (in1) -- (in2);
\draw [arrow] (in2) -- (pro1);
\end{tikzpicture}
\\ \\Above is the text cleaning pipeline that was followed. Preprocessing was a crucial step in order to increase accuracy and other metrics. The effects observed by this step are documented in the table below.

\begin{center}
\begin{tabular}{||c c c||} 
 \hline
 & Accuracy Before & Accuracy After \\ [0.5ex] 
 \hline\hline
 Gender Prediction& 70.6\% & 86.32\%  \\ 
 \hline
 Age Prediction& 67.6\% & 84.2\% \\[1ex]
 \hline
\end{tabular}
\end{center}

\section{ \\ Architecture and Pipeline}

After the text is cleaned and passed through the preprocessing pipeline, we then again pass it through the BERT preprocess module, in order to maintain a high quality of the final dataset that we feed to our network. Our next step is to prepare the dataset and feed it into our model to fine-tune BERT. A detailed step-by-step exploration of the architecture is shown below.
We used a total of three layers of BERT, and two layers of neural networks for different purposes. First, let's have a look at the BERT layers. \\

\begin{tikzpicture}[node distance=2cm]

\node (start) [startstop] {Input Layer (text input)};
\node (in1) [startstop, below of=start] {Preprocessed Text (BERT preprocess(text input))};
\node (in2) [startstop, below of=in1] {Otput Layer (BERT Encoder (Preprocessed Text))};
\draw [arrow] (start) -- (in1);
\draw [arrow] (in1) -- (in2);
\end{tikzpicture}

\hfill \break
Following are the layers of neural network that we use once the data is passed through the BERT layers. We have a total of two layers, with one being a dropout layer, which is used in order to reduce or prevent the overfitting of the model. The second layer is a sigmoid activation layer. This is required because in order to classify whether the input text is written by a male or a female, we need some probabilities, and this layer does the same. The sigmoid layer returns a number between 0 to 1. The representation of how a sigmoid curve looks and how it maps the results between 0 to 1 is as below. 
\begin{center}
    $\sigma = 1/(1+exp(-x))$
\end{center}

\pgfplotsset{compat=1.16}

\begin{tikzpicture}[declare function={sigma(\x)=1/(1+exp(-\x));
sigmap(\x)=sigma(\x)*(1-sigma(\x));}]
\begin{axis}%
[
    grid=major,     
    xmin=-6,
    xmax=6,
    axis x line=bottom,
    ytick={0,.5,1},
    ymax=1,
    axis y line=middle,
    samples=100,
    domain=-6:6,
    legend style={at={(1,0.9)}}     
]
    \addplot[blue,mark=none]   (x,{sigma(x)});
    \addplot[red,dotted,mark=none]   (x,{sigmap(x)});
    \legend{$\sigma(x)$,$\sigma'(x)$}
\end{axis}
\end{tikzpicture}

We have kept this as a convention that Male corresponds to 0, while Female corresponds to 1. So, with a number between 0 to 1, we can decide whether it maps to Male or Female.\\ 

\begin{center}
    \begin{tikzpicture}[node distance=2cm]

    \node (start) [startstop] {Dropout Layer};
    \node (in1) [startstop, below of=start] {Sigmoid Function Activation Layer};
    \draw [arrow] (start) -- (in1);
    \end{tikzpicture}
\end{center}

\hfill \break
We use the default settings of BERT, with the number of layers at 12, with 12 attention heads. The optimizer used was 'Adam', while the loss used was 'Binary Crossentropy', whose formula is given as $-{(y\log(p) + (1 - y)\log(1 - p))}$, where y is the binary indicator (0 or 1) if class label c is the correct classification for observation o, and p is the predicted probability observation o is of class c.
Finally, let's have a look at our network in its entirety to get a clear picture of what are the pathways our data will go through to get the final result.\\

\begin{tikzpicture}[node distance=2cm]

\node (start) [startstop] {Input Layer (text input)};
\node (in1) [startstop, below of=start] {Preprocessed Text (BERT preprocess(text input))};
\node (in2) [startstop, below of=in1] {Otput Layer (BERT Encoder (Preprocessed Text))};
\node (in3) [startstop, below of=in2] {Dropout Layer};
\node (in4) [startstop, below of=in3] {Sigmoid Function Activation Layer};
\draw [arrow] (start) -- (in1);
\draw [arrow] (in1) -- (in2);
\draw [arrow] (in2) -- (in3);
\draw [arrow] (in3) -- (in4);
\end{tikzpicture} 

\section{\\Results and Conclusion}
After feeding our data in the above pipeline, we get some interesting results. The training was done for 150 epochs on a TensorFlow-based kernel. First, we'll discuss about the results we got while trying to predict age from the given text. Let's have a look at how the textual inputs were classified by our model.
\begin{figure}[htp]
    \centering
    \includegraphics[width=8cm]{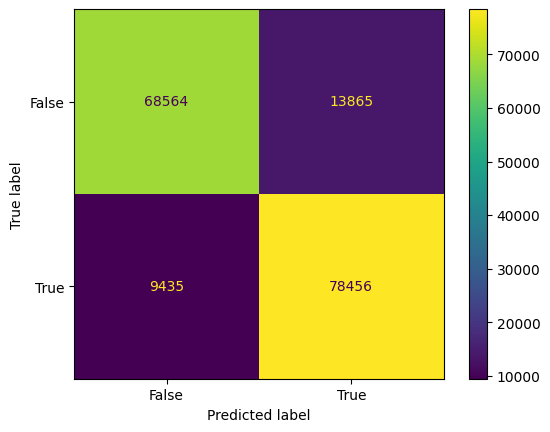}
    \caption{Confusion matrix for gender prediction}
    \label{fig:galaxy}
\end{figure}

\begin{center}
\begin{tabular}{||c c||} 
 \hline
 True Positive & 78456 \\ [1ex] 
 \hline
 True Negative & 68564  \\ 
 \hline
 False Positive & 13865 \\
 \hline
 False Negative & 9435 \\[1ex] 
 \hline
\end{tabular}
\end{center}
Using the above table, we can see the number of true positives, false positives, true negatives, and false negatives. Using these values, we can find some more useful metrics which can further inform us about how well our model has performed or how badly has it erred. If we try to evaluate our model on the basis of some standard metrics, the result is as follows: 
\hfill \break
\begin{center}
\begin{tabular}{||c c c c||} 
 \hline
 Precision & Recall & Accuracy & F1 Score \\ [1ex] 
 \hline \hline
 0.8498 & 0.8926& 0.8632&0.8706 \\[1ex] 
 \hline
\end{tabular}
\end{center}
Now, when we focus on the prediction class of age, then the results that we have observed are as follows.
\begin{figure}[htp]
    \centering
    \includegraphics[width=8cm]{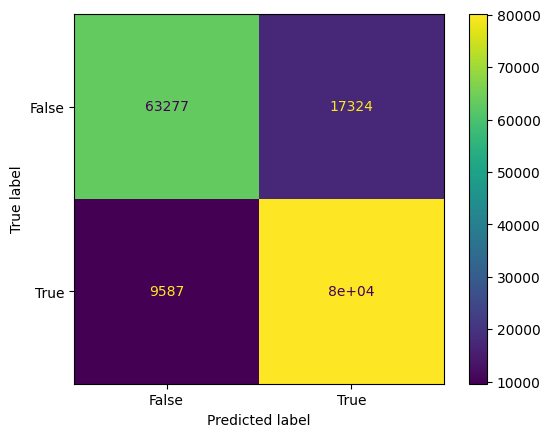}
    \caption{Confusion matrix for age prediction}
    \label{fig:galaxy}
\end{figure}

\begin{center}
\begin{tabular}{||c c||} 
 \hline
 True Positive & 80132 \\ [1ex] 
 \hline
 True Negative & 63277  \\ 
 \hline
 False Positive & 17324 \\
 \hline
 False Negative & 9587 \\[1ex] 
 \hline
\end{tabular}
\end{center}
if we have a look at the metrics, then we can also compare the two, age and gender prediction among themselves and find out for which task our algorithm does a better task.
\hfill \break
\begin{center}
\begin{tabular}{||c c c c||} 
 \hline
 Precision & Recall & Accuracy & F1 Score \\ [1ex] 
 \hline \hline
 0.8223 & 0.8932& 0.8420&0.8562 \\[1ex]
  \hline
\end{tabular}
\end{center}
 \hfill \break
 The results we got are pretty good, especially if we consider the previous works\cite{Goswami_Sarkar_Rustagi_2009}. The great thing is that we have a healthily good score of both precision and recall and an equally good F1 score. If we consider a comparison between the prediction of age class and gender, then the results show that our algorithm does a better job in predicting gender in comparison to predicting the age class, or group. This may be attributed to the fact that gender data had two labels, while the age class had 3 labels (10s, 20s, and 30s).
 \hfill \break
\begin{center}
\begin{tabular}{||c c c c c||} 
 \hline
 Feature & Precision & Recall & Accuracy & F1 Score \\ [1ex] 
 \hline \hline
 Age & 0.8223 & 0.8932& 0.8420&0.8562 \\
 Gender & 0.8498 & 0.8926& 0.8632&0.8706 \\[1ex]
  \hline
\end{tabular}
\end{center}
\hfill \break
The prediction of gender has better precision, accuracy, and F1 score, while it lags marginally in recall in comparison to the prediction of age class.\\
We can conclude that BERT-based textual classification\cite{10.1007/978-3-030-32381-3_16} works really well for age and gender prediction tasks, and can be used in many real-case scenarios. We also observe that this method is marginally better for gender prediction, or in general we can say that this method works well for classification if the dataset has a less number of classes.

\bibliographystyle{splncs04}

%
%
%

\end{document}